\documentclass{article}

\usepackage{spconf,amsmath,graphicx}
\usepackage{mathptmx} 
\usepackage{times} 

\usepackage{multirow}
\usepackage{textcomp}
\usepackage{colortbl}
\usepackage{xcolor}
\usepackage{array,tabularx}
\usepackage{enumitem}
\usepackage{adjustbox}

\title{A Survey On 3D Inner Structure Prediction from its Outer Shape }

\name{Mohamed Mejri$^{1}$, Antoine Richard $^{2}$, C\'edric Pradalier$^{3}$  }
\address{ \small School of Electrical and Computer Engineering, Georgia Institute of Technology, Atlanta, USA$^{1,2}$\\\small UMI2958 GT-CNRS, Metz, France$^{3}$}

\begin{document}
\maketitle

\begin{abstract}
    The analysis of the internal structure of trees is highly important for both forest experts, biological scientists, and the wood industry. Traditionally, CT-scanners are considered as the most efficient way to get an accurate inner representation of the tree. However, this method requires an important investment and reduces the cost-effectiveness of this operation. Our goal is to design neural-network-based methods to predict the internal density of the tree from its external bark shape. This paper compares different image-to-image(2D), volume-to-volume(3D) and Convolutional Long Short Term Memory based neural network architectures in the context of the prediction of the defect distribution inside trees from their external bark shape. Those models are trained on a synthetic dataset of 1800 CT-scanned look-like volumetric structures of the internal density of the trees and their corresponding external surface.    
\end{abstract}


\section{INTRODUCTION}

     For the wood industry, several studies\cite{4}\cite{5}\cite{6} showed that the transformation of the wood based on the internal density of the log improved the value recovery. Knots are considered as the main important inner element of the wood. They are either characterized manually or through X-Ray CT-scanning.\\
     As an example, using X-Ray information, in a study based on several hundreds of logs of Scots pine
     (Pinus sylvestris L.) and Norway spruce (Picea abies L.), \cite{5} assessed an average increase of the recovery value by $13\%$ when compared to a sawing position of the log based on the outer shape. However, the analysis of the inner structure of the log requires highly experienced forests experts and an expensive investment ($\approx$5M\$).\\
     From a living branch until a knot in duramen, these different stages have a more or less evident impact on the external bark roughness.\\
      Those arguments prove the existence of a potential correlation between the internal density of the log and its outer shape. To evaluate this correlation, we propose to compare different 2D and 3D based neural network architectures. However, finding complex defects in the inner structure of a log from its internal shape using deep learning methods requires a dataset with thousands of real CT-scanned log with their corresponding tree bark shape. Hence, a synthetic CT-scanned look-like dataset is generated. 
      To predict the internal density of the tree, three main variety of neural network architectures were tested, which are 2D Encoder-Decoder architectures, 3D Encoder-Decoder models, and Convolutional-LSTM (CLSTM) based Encoder-Decoder. 
     
      \begin{figure}[!tb]
        \centering
        \includegraphics[width = .5\columnwidth]{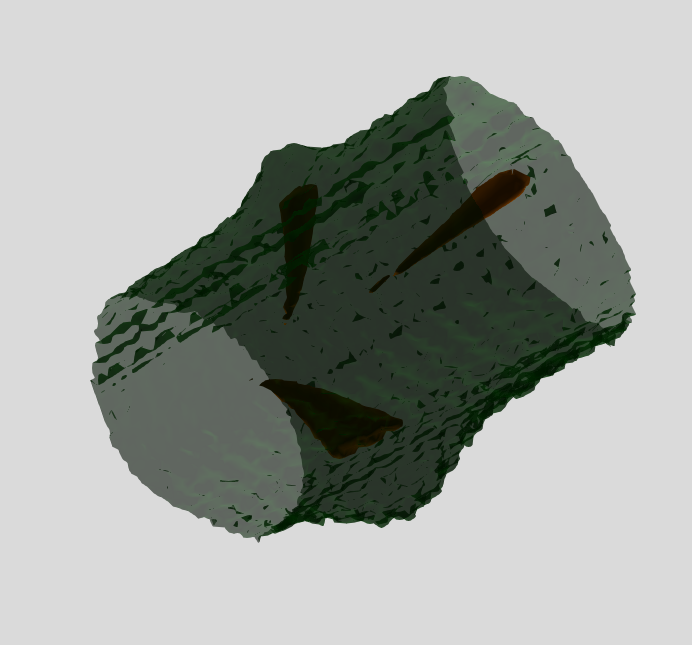}
        \caption{\small(Green) surface (Brown) iso-surface of inner density}
    \end{figure}

\section{RELATED WORK}
In pixel-wise regression, the typical structure of models is based on an encoder followed by a decoder composed mainly of convolutional layers. In this paper, we will focus on 2-D, 3-D, and CLSTM based encoder-decoders.

2D Encoder-Decoder architectures such as SegNet\cite{badrinarayanan2015segnet}, U-Net\cite{ronneberger2015unet} take an image as an input, and, when used for pixel-wise regression, output a matrix of floats of same height and width as the input. These architecture are comprised of a contractive path (the encoder) and an expansive path (the decoder). 

Fully Convolutional Networks such as DeepLab\cite{chen2018encoderdecoder}, FCN\cite{long2014fully}, and  PSPNet\cite{zhao2016pyramid} shows almost the same results but the decoding process is slightly different. The decoding is built using a spatial pooling pyramid, instead of the expansive path. This results in higher accuracy in pixel-wise classification but not in pixel-wise regression. Indeed, due to the upsampling mechanism used in the Spatial Pyramidal Pooling, the regression results are not as good as SegNet's or U-Net's results.

Since the development of 3D data acquisition techniques, especially in medical imaging, several 3D neural network architectures were designed to perform volumetric segmentation. 3D-U-Net\cite{iek20163d} was designed to perform kidney\cite{iek20163d} segmentation. It extends the U-Net architecture \cite{ronneberger2015unet}, by replacing all 2D operations with their 3D counterparts.
\cite{chen2016voxresnet} introduced a voxel-wise residual neural network based on residual neural network\cite{he2015deep}. It consists of three stacked residual modules followed by four 3D-deconvolutional layers. According to \cite{chen2016voxresnet}, VoxResNet achieves better results  than 3D-U-Net \cite{iek20163d} after being tested on MICCAI MRBrainS challenge data \cite{chen2016voxresnet}.

Both 3D and 2D encoder-decoder architectures discussed so far cannot capture the correlation that may exist between the successive elements in the dataset. Recurrent Neural Network and, more specifically, Long Short Term Memory (LSTM) structures are widely used in natural language processing to achieve sequence-to-sequence processing.
Those neural networks aim to retrieve correlations between words in the same sentence. \cite{nabavi2018future} introduces Convolutional-LSTMs (CLSTM)  
\cite{shi2015convolutional}) based encoder-decoder structure.\\ The encoder structure generates feature-maps from the input images that are fed to the CLSTM\cite{shi2015convolutional} module. It consists of a Long Short Term Memory module where the weights are replaced with a filter bank of a convolutional layer. The decoder is composed of several deconvolutional\cite{dumoulin2016guide} layers and combines the outputs of different CLSTM modules and generates the segmentation map for the next time-step. 
Bidirectional CLSTM based segmentation structures have also been introduced and aim to
capture the temporal information in both directions.

\section{METHOD}
This section details the different strategies and approaches used to predict the internal structure of the tree from its external shape. 
\subsection{2D neural network architectures}
      \begin{figure}[thpb]
        \centering
        \includegraphics[width = \linewidth,height=4cm]{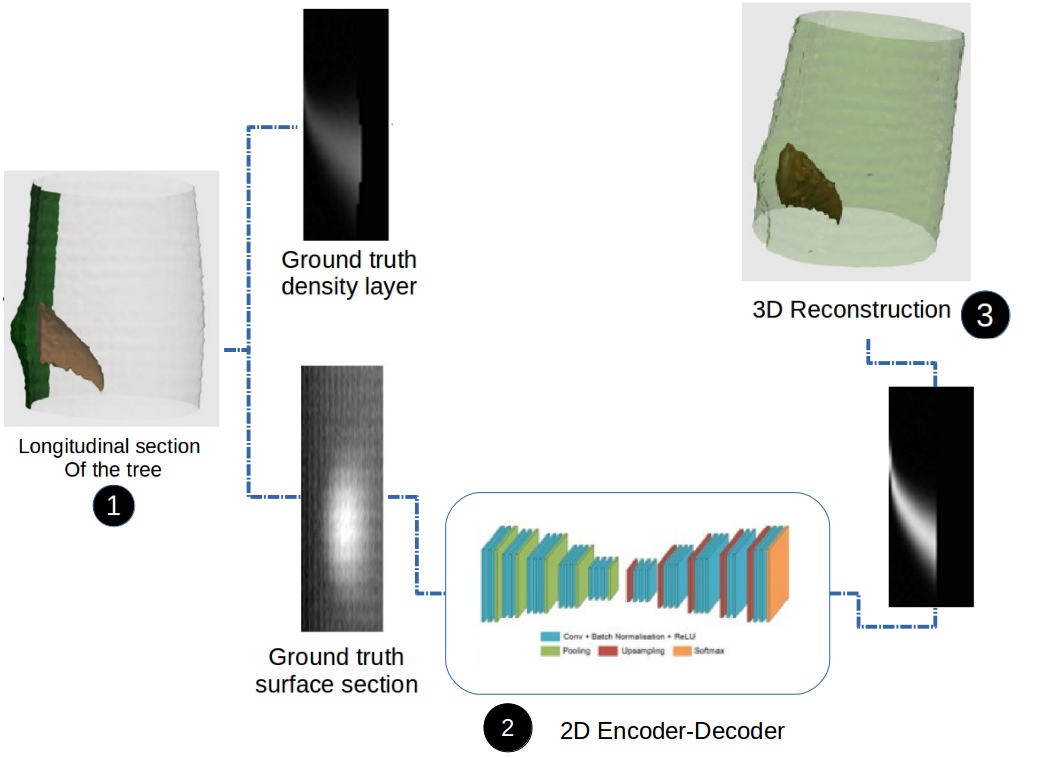}
        \caption{\small Different steps for 2D prediction and 3D reconstruction}
        \label{fig:2d2d}
    \end{figure}
    
The first strategy relies on predicting a slice of the density $d = f(r,\theta=\theta_{0},z)$ within the radial longitudinal plane using its correspondent surface $s=S(\theta \in [\theta_{0}-\frac{\pi}{12},\theta_{0}+\frac{\pi}{12}],z)$ using auto-encoders like architectures. Fig \ref{fig:2d2d} illustrates the prediction pipeline used in image-to-image predictions.    
 As stated in the previous section, three variety of 2D-Encoder decoder will be used: SegNet \cite{badrinarayanan2015segnet}, U-Net \cite{ronneberger2015unet}, and DeepLabV3+ \cite{chen2018encoderdecoder}.
To use SegNet\cite{badrinarayanan2015segnet} architecture as a pixel-wise regression encoder-decoder, the loss function has to be modified. We used an $L2$ loss(mean-squared error). Furthermore, we changed the depth of the encoder and the decoder of the original SegNet \cite{badrinarayanan2015segnet} due to the low-resolution input to prevent the model from overfitting. We removed the last and the first convolutional block respectively from the encoder and the decoder architectures. Additionnaly, we replaced the rectified linear function (ReLU) with parametric-rectified linear function $P-ReLU$\cite{he2015delving} which is a learned-slope value version of leaky-ReLU. It improves model fitting while avoiding over-fitting issues. SegNet is known to be a very Deep-Network, often in Deep Networks, part of the information is lost due to the large amount of operation the data has to go through. This, often results in degraded performance. 
Hence, we tested other encoder-decoder architectures that feature skip-connections, that alleviate the very-deep networks issues.
U-Net\cite{ronneberger2015unet} architecture is one of them.Due to low-resolution constraints, the original contracting and expanding path of the U-Net\cite{ronneberger2015unet} architecture were reduced (i.e., the 1024 filters convolution layer was removed).
The DeepLabV3+ \cite{chen2018encoderdecoder}is a revisited version of DeepLab\cite{chen2016deeplab} architecture containing the following structures: Atrous Spatial Pooling Pyramids, Xception blocks,Depth-wise separable convolutions. 
In the original implementation of DeepLabV3+ \cite{chen2018encoderdecoder}, the L2 regularization strategy aims to avoid overfitting  with a weight decay of $4e-5$. Due to instability issue while using DeepLabV3+ \cite{chen2018encoderdecoder} for regression, we decided to add dropout at the end of all convolutional layers instead of performing L2 regularization.

\subsection{3D neural network architectures}
In this section, we will represent the internal density of the log as an entire volumetric structure.  \\Three different variety of deep learning models will be tested: 3D-U-Net \cite{iek20163d}, 3D-SegNet and VoxResNet \cite{chen2016voxresnet}. 
3D-SegNet is inspired by 2D-SegNet \cite{badrinarayanan2015segnet}, by replacing the 2D layers with their 3D counterpart and removing convolutional layers in the encoder and the decoder to reduce the memory, and time expenses, and prevent overfitting. 
It consists of a transformed VGG-16 \cite{simonyan2014deep} based encoder with 10 3D Convolutional layers. The last three 512 Convolutional layers have been removed. Each encoder layer has a corresponding decoder layer, and hence the decoder network has ten layers. A Batch-normalization layer, and a Parametric-ReLU activation layer follows each convolutional layer. 3D-max-pooling layer with 2x2x2 strides to avoid overlapping is used to down-sample the 3D feature map. Similarly, in the decoder 3D-Up-sampling layer performs 2x2x2 up-sampling while keeping max-pooled voxels at the same positions. To prevent overfitting, a dropout of 0.1 is added at the end of each convolutional layer.  

We kept the original structure of 3D-U-Net \cite{iek20163d} but made some changes in the loss function and overfitting strategy. The original 3D-U-Net \cite{iek20163d} is trained on partially annotated voxels. Hence, the loss function has to be weighted with zero weight when the data is not annotated and one otherwise. In our case, we are performing regression with a fully-annotated dataset; we used unweighted mean-squared as a loss function. Furthermore, to prevent overfitting, a dropout of 0.1 was added after each 3D convolutional layer. 

One advantage of VoxResNet \cite{chen2016voxresnet} over 3D-U-Net \cite{iek20163d}  is the ability to build deeper encoder and decoder networks while avoiding overfitting. We didn't make a significant change in the original architecture. We removed the final classification layer and kept the multi-level contextual information consisting of 4 3D-deconvolutional layers with respectively 1,2,4 and 8 degrees of strides. Dropout of 0.1 is added after each convolutional layer to prevent overfitting.
\subsection{CLSTM based network architectures}
 We modified the original architecture for time and memory expenses purposes. The SegNet \cite{badrinarayanan2015segnet} structure is used to build the encoder and the decoder of this model. To capture the correlation between the different cross-sections of the internal density of the log while reducing the computation costs, we only keep the CLSTM\cite{shi2015convolutional} layer at the end of the encoder (bottleneck). To avoid overfitting, a dropout of 0.1 was added after each convolutional layer and inside the CLSTM\cite{shi2015convolutional} Layer. Practically, handling chronologically ordered images is challenging. To overcome this problem, we used a wrapper called TimeDistributed Layer which enables applying convolutional operations multiple time to multiple input time steps and hence provides a sequence of feature maps to the LSTM model to work on.
Adam optimizer is used 
with a learning rate set to $10^{-3}$.  

Intuitively the correlation between a successive cross-section of log exists in forward and backward directions. To capture the correlation in both directions, The bidirectional CLSTM version of SegNet \cite{badrinarayanan2015segnet} could be useful. It has almost the same architecture as the original CLSTM based SegNet \cite{badrinarayanan2015segnet}.The main difference consists of a Bidirectional layer that wraps the old CLSTM layer. However, to prevent overfitting, we decided to increase the dropout from 0.1 to 0.2 inside the CLSTM layer.  
\section{EXPERIMENTS}

\subsection{dataset}
The dataset is composed of 1800 low-resolution synthetic logs: training neural network architectures on a low-resolution image is less memory-consuming and doesn't impact the quality of prediction.\\
The dataset is balanced in terms of the number of logs per branch, it is composed of 300 k-branches logs where k varies between 2 and 7.        
We chose to represent the tree surface as a function $s(r,\theta,z)$ where $\theta$ and $z$ refer to the polar coordinate of the tree. At the same time, $r$ indicates the variation of the radius of the cross-section of a tree log compared to a fixed radius ($r=0.5$) cylinder.
 The texture and the main transverse shape of the tree were modeled respectively as a high and low-frequency cosine. The longitudinal form of the log is modeled as a decreasing function starting from its basis. Furthermore, The branches are modeled as a 2D-Gaussian function.

We modeled the internal density of the log as a function $d(r,\theta,z)$ . The branch, a region with high density, is modeled as a combination of square root functions and linear functions. The figure below illustrates two different sections of a log as well as two external surfaces projection.   
      \begin{figure}[thpb]
        \centering
        \includegraphics[width = 0.8\linewidth,height=2cm]{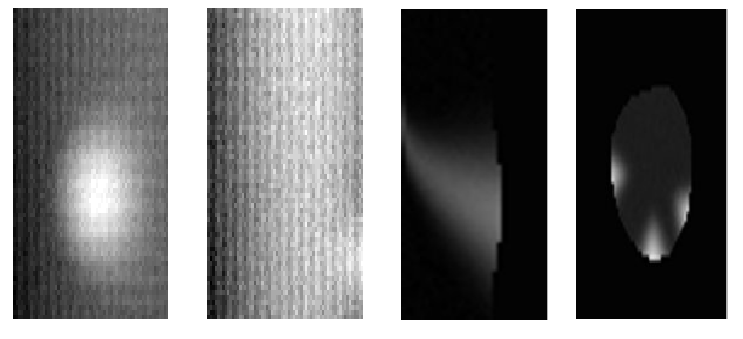}
        \caption{\small From left to right: surface with a branch, surface without branch, longitudinal section of a log, cross section of a log }
    \end{figure}


\subsection{Network Training and validation }
We split the original dataset into three subsets: training set, validation set, and testing set with the respective proportion $80\%$, $4\%$, and $16\%$.\\
For computational constraints, the batch size is fixed to two volumes, 100 images, and a sequence of 64 images for respectively 3D models, 2D models, and CLSTM based models. Each model is trained for 50 epochs.  At the end of each epoch, a validation test is performed, and we save only the weights of the model with a minimum validation error. All experiments were carried out on an IBM Power Systems AC922 with 256 GB of RAM and 4 NVIDIA V100 16 GB GPGPU (using only a single GPGPU). 
 
\section{RESULTS}


\subsection{Architectures}
    In this section, we will assess the prediction of 280 logs with their corresponding external surface. To assess the training and validation error, we use a mean-squared error. For the testing process, we used the root mean squared error (RMSE).
    For the 2D models, there are two possible ways to make the assessment either with the predicted density layers or with the reconstructed volumetric structures. Practically, there is no significant difference between the two approaches. 
    The table I shows the mean performance of the different 2D,3D, and CLSTM based neural network architectures when tested on 40 various k-branches logs ($k\in \{2,5,7\}$)\\
    
\begin{table}[h]
\caption{Best results for each type of architecture }
\label{tab:ARCHjoint}
\begin{center}
\begin{adjustbox}{center, width=0.75\columnwidth} 
\begin{tabular}{|l||c|c|c|c|}

\hline
\multirow{2}{*}{Architecture} & \multicolumn{3}{c|}{RMSE $(10^{-2})$}     & \multirow{2}{*}{Parameters} \\ \cline{2-4}
                              & 2 $Branches$ & 5 $Branches$ & 7 $Branches$ &                             \\ \hline\hline
SegNet                        & \bf1.27       & \bf1.66       & 2.46       & 34 M                        \\ \hline
U-Net                         & 1.33       & 1.68       & \bf2.42       & 36 M                        \\ \hline
DeepLabV3+                    & 3.17       & 3.23       & 3.48       & 42 M                        \\ \hline\hline
3D-SegNet                     & 2.39       & 3.13       & 3.92       & 144 M                       \\ \hline
3D-U-Net                      & 2.48       & 3.10       & 3.70       & 114 M                       \\ \hline
VoxResNet                     & 2.86       & 3.37       & 3.92       & 35 M                        \\ \hline\hline
CLSTM-SegNet                  & 2.93       & 4.38       & 5.03       & 31 M                        \\ \hline
Bidir-CLSTM-SegNet            & 2.49       & 3.4        & 4.23       & 52 M                        \\ \hline

\end{tabular}}
\end{adjustbox}
\end{center}
\end{table}

We conclude that SegNet \cite{badrinarayanan2015segnet} achieves better prediction results than DeepLabV3+ and U-Net for two branches logs. In contrast, U-Net \cite{ronneberger2015unet} achieves better results for predicting the internal density of logs with a high number of branches. Those results are expected because, unlike the SegNet \cite{badrinarayanan2015segnet} model, U-Net \cite{ronneberger2015unet} architecture focus more on capturing the information lost through the encoder structure, which may be relevant when it comes to more complex log structure (i.e., logs with a high number of branches). DeepLabV3+ \cite{chen2018encoderdecoder} fails compared to the other architectures to achieve good results: The DeepLabV3+ \cite{chen2018encoderdecoder} architecture is significantly complex compared to SegNet \cite{badrinarayanan2015segnet} and U-Net \cite{ronneberger2015unet} (42M parameters for DeepLabV3+\cite{chen2018encoderdecoder}) which may lead to an overfitting. Furthermore, the upsampling and downsampling structures of the DeepLabV3+\cite{chen2018encoderdecoder} fail to guarantee a better density layer reconstruction. \\3D-SegNet and 3D-U-Net \cite{iek20163d} achieve better performance than VoxResNet \cite{chen2016voxresnet}. The 3D-SegNet model keeps the same indexes of maximum voxels value after upsampling. Unlike 3D-U-Net \cite{iek20163d} and VoxResNet \cite{chen2016voxresnet}, 3D-SegNet doesn't contain real skip-connection or concatenation layer and hence doesn't recover information lost at the end of the encoder. This can be crucial when it comes to a more complicated structure, i.e., log with a high number of branches.   

To capture the correlation between successive cross-sections of the same log, we tested two versions of the CLSTM based SegNet. Intuitively, building a model that considers a bidirectional correlation between consecutive cross-sections of a log is more relevant than observing the correlation between the different slices of the log in one single direction. This intuition is confirmed according to the RMSE results shown above. According to the table I, 2D-models achieves the least RMSE scores while reducing the memory expenses.  

\subsection{3D visual results }
We used  $Paraview^{\copyright}$, an open-source multi-platform application for 3D visualization, to assess the 3D rendering of the predicted iso-surface of the internal density of the tree. The figures below show the predicted iso-surface with different models for the 6-branches log. Based on the RMSE evaluation, we chose to assess only the 2D, 3D, and CLSTM based models with the highest performance. We conclude that the 2-D models achieve better results than 3-D models in term of iso-surface correspondence between the ground truth (Green volume) and the predicted log (Red volume).    
      \begin{figure}[thpb]
        \centering
        \includegraphics[width = \linewidth,height=3cm]{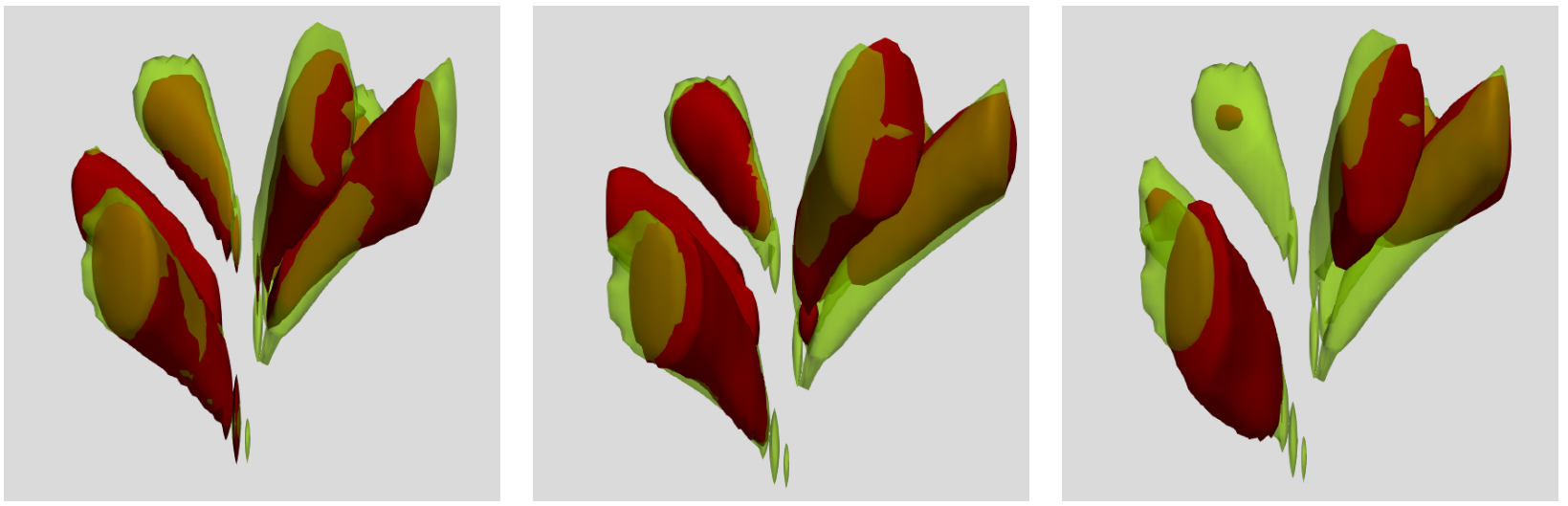}
        \caption{\small From left to right: 3D-U-Net, U-Net, Bidirectional CLSTM based SegNet}
        \label{fig:rnn}
    \end{figure}

\section{CONCLUSIONS}
In this paper, we compared different neural network architectures to predict the inner density of synthetically made logs from their outer surface shape. Unlike 3D and CLSTM based models that are too complicated in terms of parameters number, 2D models are very simple encoder-decoder architectures with commonly used layers (Conv2D, Max-pooling2D,etc.). Thus they are more robust to overfitting. They achieve better performance in terms of RMSE evaluation and visual quality assessment of the iso-surfaces than 3D-based models and CLSTM-based models. However, integrating the correlation assessment in density layer prediction (2D-models) with recurrent neural network approaches could give a better result and will be explored. Further work should be focus on the prediction of real CT-scanned trees dataset where the defect distribution is more random and less correlated to the external surface of the tree bark.  
\section{ACKNOWLEDGEMENT}
We would like to thank Lionel Clavien from our partner Inno-
Boost SA in Switzerland for providing us access to the IBM
Power  Systems  AC922  server  used  for  our  experiments  as
well as some starting help on the platform




\bibliographystyle{IEEEbib}
\bibliography{references}

\end{document}